%% file: 0-main.tex
\pdfoutput=1

\documentclass[letterpaper, 10 pt, conference]{ieeeconf}  

\IEEEoverridecommandlockouts                              

\usepackage[backend=bibtex,
            hyperref=true,
            url=false,
            isbn=false,
            doi=false,
            backref=false,
            style=ieee,
            natbib=true,
            mincitenames=1,
            maxcitenames=1,
            citestyle=numeric-comp,
            sorting=nyt,
            block=none]{biblatex}

\input{0-preamble.tex}

\input{0-defs.tex}

\title{\LARGE \bf
LEGS: Learning Efficient Grasp Sets for Exploratory Grasping}

\author{Letian Fu$^{1}$, Michael Danielczuk$^{1}$, Ashwin Balakrishna$^{1}$, Daniel S. Brown$^{1}$,\\ Jeffrey Ichnowski$^{1}$, Eugen Solowjow$^{2}$, Ken Goldberg$^{1}$%
\thanks{$^{1}$The AUTOLab at UC Berkeley, $^{2}$Siemens Research Lab}
}

\begin{document}

\maketitle

\begin{abstract}
While deep learning has enabled significant progress in designing general purpose robot grasping systems, there remain objects which still pose challenges for these systems. Recent work on Exploratory Grasping has formalized the problem of systematically exploring grasps on these adversarial objects and explored a multi-armed bandit model for identifying high-quality grasps on each object stable pose. However, these systems are still limited to exploring a small number or grasps on each object. We present \algname (\algabbr), an algorithm that efficiently explores thousands of possible grasps by maintaining small active sets of promising grasps and determining when it can stop exploring the object with high confidence. Experiments suggest that \algabbr can identify a high-quality grasp more efficiently than prior algorithms which do not use active sets. In simulation experiments, we measure the gap between the success probability of the best grasp identified by \algabbr, baselines, and the most-robust grasp (verified ground truth). After 3000 exploration steps, \algabbr outperforms baseline algorithms on 10/14 and 25/39 objects on the Dex-Net Adversarial and EGAD! datasets respectively. We then evaluate \algabbr in physical experiments; trials on 3 challenging objects suggest that \algabbr converges to high-performing grasps significantly faster than baselines. See \url{https://sites.google.com/view/legs-exp-grasping} for supplemental material and videos.

\end{abstract}

\input{1-introduction.tex}
\input{2-related-work.tex}

\input{3-problem-statement.tex}
\input{3.5-prelims.tex}
\input{4-methods.tex}

\input{5-experiment-and-result.tex}
\input{6-physical-experiments.tex}
\input{7-summary.tex}

\input{8-acknowledgement.tex}

\renewcommand*{\bibfont}{\footnotesize}
\printbibliography 
\clearpage
\input{9-appendix}

\end{document}

%% file: 0-preamble.tex
\usepackage{graphics}
\usepackage[pdftex]{graphicx}
\usepackage{xcolor}
\usepackage{wrapfig}
\DeclareGraphicsExtensions{.pdf,.png,.jpg}
\usepackage{epsfig}
\usepackage[font={small}]{caption}
\usepackage{subcaption}
\usepackage[rightcaption]{sidecap}
\usepackage{pbox}
\usepackage{makecell}

\usepackage{bigstrut}
\usepackage{mathtools}
\usepackage{amssymb,amsmath} 
\usepackage{gensymb} 
\usepackage{nicefrac}       
\numberwithin{equation}{section} 

\DeclareMathAlphabet{\mathcal}{OMS}{lmsy}{m}{n}
\DeclareSymbolFont{largesymbols}{OMX}{cmex}{m}{n}
\usepackage{textcomp} 
\usepackage[linesnumbered, ruled,vlined,noend]{algorithm2e}

\usepackage{array} 
\usepackage{tabularx}
\usepackage{multirow}
\usepackage{multicol}
\usepackage{booktabs}
\usepackage{tabulary}

\usepackage[utf8]{inputenc}
\usepackage{units}
\usepackage{bm}
\usepackage{xspace}
\usepackage{flushend}
\usepackage{balance} 
\usepackage{csquotes}
\usepackage{makeidx}
\usepackage{blindtext}

\usepackage{autolabtools}

\let\labelindent\relax
\usepackage{enumitem}

\usepackage{tikz}
\usetikzlibrary{backgrounds}

\usepackage{ragged2e}
\usepackage{soul} 
\usepackage{subfiles} 

\usepackage[protrusion=true,expansion=true]{microtype}
\usepackage{url}
\makeatletter
\g@addto@macro{\UrlBreaks}{\UrlOrds}
\makeatother

\usepackage{hyperref}
\hypersetup{
    colorlinks=true,
    linkcolor=black,
    citecolor=black,
    filecolor=cyan,
    urlcolor=black
}

\usepackage{siunitx}

\renewcommand{\baselinestretch}{1.0}




%% file: 0-defs.tex
\newcommand{\algname}{Learned Efficient Grasp Sets\xspace}
\newcommand{\algabbr}{LEGS\xspace}

%% file: 1-introduction.tex
\section{Introduction}
\label{sec:introduction}





Recent advances in deep learning have enabled the development of universal grasping systems that can robustly grasp a wide variety of objects ~\cite{ggcnn, Mahlereaau4984, viereck2017learning, pinto2016supersizing, mahler2017dex, mahler2017binpicking}. 
However, these systems can still struggle to grasp objects with adversarial~\cite{morrison2020egad, Wang2019AdversarialGO} geometries or which are significantly out of distribution from the objects seen during training. This problem is common in many industrial settings, in which newly manufactured machine parts for custom applications may look very different from the objects in the datasets typically used for training universal grasping systems.

Recently, bandit-style algorithms have been used to augment general-purpose grasping policies by rapidly adapting them to specific objects~\cite{laskey2015, lu2020, eppner2017, li2020}. Recently, \citet{danielczuk2020exploratory} introduced Exploratory Grasping, where a robot learns to grasp novel objects through online exploration of grasps and stable poses. Their algorithm, Bandits for Online Rapid Grasp Exploration Strategy (BORGES), learns robust pose-specific grasping policies. 
However, BORGES limits exploration to a fixed set of 100 grasps per stable pose, possibly to overlooking other high-quality grasps.

In this work, we extend \citet{danielczuk2020exploratory} to explore thousands of grasps per stable pose. Considering grasp sets of this scale increases the likelihood of converging to a robust grasp, but also makes efficient exploration challenging. To address this challenge, we propose \emph{\algname{}} (\algabbr), which adaptively curates an active set of promising grasps rather than restricting exploration to a small fixed subset. The key insight is to use a combination of priors from a universal grasping system and online trials to maintain confidence bounds on grasp-success probabilities. \algabbr{} uses these bounds to (1) update the grasps in its active set and (2) decide when to stop exploring.
\begin{figure}[t] 
    \centering
    \includegraphics[width=\linewidth]{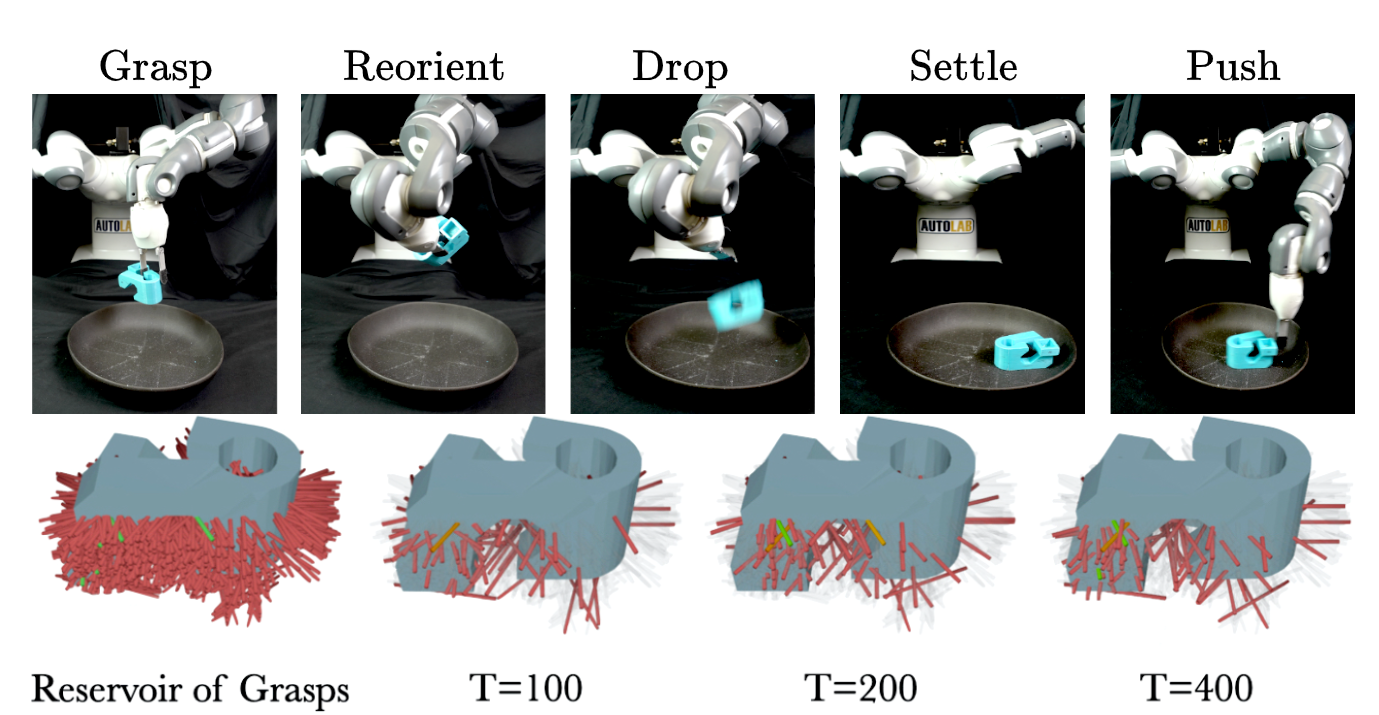}
    \caption{\textbf{Top: \algabbr in Physical Experiments: } \algabbr repeatedly attempts grasps on an object, and if the grasps are successful, it re-drops the object into a new stable pose. \textbf{Bottom: \algabbr Active Set Evolution: } \algabbr works by adaptively curating a small active set of promising grasps out of a large reservoir of grasp candidates (left). As exploration progresses, \algabbr refines its active set (shown in bolded red/green) to contain higher quality grasps (right).}
    \label{fig:splash}
\end{figure}


This paper makes the following contributions: (1) a novel adaptive multi-armed bandits algorithm that curates a small set of high-performing grasps by actively removing and resampling grasps based on performance bounds and a novel termination condition that enables a robot to predict (with high confidence) when it reaches a desired level of performance;
(2) a \textit{self-supervised} physical grasping system where a robot explores candidate grasps with minimal human intervention (roughly 1 in every 100 grasp attempts); (3) simulation and physical experiments suggesting that \algabbr can identify higher quality grasps within a fixed time horizon than prior algorithms which do not learn an active set.


%% file: 2-related-work.tex
\section{Related Work}
\label{sec:related_works}




\subsection{Universal Grasping Algorithms}
Recent robotic grasping algorithms generalize to a wide range of objects~\cite{grasping-survey}. Open-loop algorithms synthesize grasps and predict their quality based on the geometry of the object, and then plan and execute a motion to attempt a high-quality grasp without feedback~\cite{lenz2015, mahler2017dex, pinto2016supersizing, mahler2017binpicking, Mahlereaau4984}. Closed-loop grasp planners that use vision-based gripper servoing~\cite{ggcnn, viereck2017learning} and RL~\cite{qtopt, mtopt} have also been popular in prior work. 
\algabbr is designed to leverage priors from these universal grasping algorithms to efficiently learn a robust grasp policy for a specific, difficult-to-grasp object~\cite{morrison2020egad, Wang2019AdversarialGO}. We use priors from Dex-Net 4.0~\cite{Mahlereaau4984}, a general grasp planner that learns a grasp-quality estimator from a large dataset of 3D object models in simulation and then uses this estimator to sample and evaluate the quality of grasps in physical trials. 
\subsection{Multi-Armed Bandits}
Prior work on multi-armed bandits~\cite{slivkins2019introduction} 
has studied settings where the number of actions is large compared to the number of timesteps allocated for exploration~\cite{teytaud:inria-00173263, wang2017, jedor2021greedy, berry1997, Wang2008AlgorithmsFI, Heide2021BanditsWM, Carpentier2015SimpleRF}. 
One popular algorithmic framework for this setting is called \textit{best arm identification}, where the goal is to adaptively reject a set of arms from consideration when there is high confidence that they are suboptimal~\cite{sr10, karnin2013, multi2013}. \algabbr builds on these ideas, by adaptively filtering actions from an active set by maintaining confidence bounds on the reward corresponding to each action. This mechanism makes it possible to efficiently perform best arm identification across multiple bandits problems, where each bandit problem represents a distinct stable pose of an object. \algabbr can quickly converge to high-quality grasps on problems with thousands of grasps per stable pose.

\subsection{Exploratory Grasping}
Universal grasping algorithms often struggle with certain objects~\cite{Wang2019AdversarialGO, morrison2020egad}.
\citet{danielczuk2020exploratory} show that grasping algorithms such as Dex-Net~\cite{Mahlereaau4984} are difficult to fine-tune online on such objects, and propose \textit{Exploratory Grasping}, a problem formulation where the objective is to perform rapid online adaptation to grasp specific, unknown objects. To achieve this, prior works sample a fixed set of grasps on specific object stable poses and apply multi-armed bandit algorithms to rapidly identify high-performing candidates~\cite{laskey2015, lu2020, eppner2017, li2020}. \citet{danielczuk2020exploratory} extend these ideas with 
BORGES, which explores grasps across all object stable poses by using Thompson sampling and a learned Dex-Net prior~\cite{li2020}. 
However, BORGES can often overlook high-quality grasps since it restricts exploration to a small initial set of grasps. To address this issue, \algabbr begins with a large set of grasp candidates and adaptively curates sets of promising grasps by adding and removing grasp candidates during exploration. By doing this, \algabbr is able to converge to better long-term performance than BORGES (which uses a small fixed set of grasps), while also learning to robustly grasp an object faster than baselines that seek to directly explore large sets of grasp candidates.

%% file: 3-problem-statement.tex
\section{Problem Statement}
\label{sec:prob_form}

\textbf{Overview:}
Given a difficult-to-grasp polyhedral object of unknown geometry that rests on a planar surface and is viewed by an overhead depth camera, we seek to learn to successfully grasp the object in all of its stable poses. 

\textbf{Problem Setup:}
Given a polyhedral object $o$, let $N$ be its number of stable poses. Each stable pose $s \in \{1, 2, \hdots N\}$ is associated with a landing probability $\lambda_s$, which indicates the probability of the object landing in pose $s$ when released from sufficient height in a randomized orientation~\cite{795790, moll2002}. 
Following \citet{danielczuk2020exploratory}, we model our problem as a finite-horizon Markov Decision Process $\mathcal{M} = (\mathcal{S}, \mathcal{A}, T, R, H)$.
We let $\mathcal{S}$ be the set of equivalence classes of distinguishable stable poses of the object and $\mathcal{A}$ be the set of all possible grasps on the object. 
Thus, $\mathcal{A} = \bigcup_{s \in \mathcal{S}}^N \mathcal{A}_s$, where $\mathcal{A}_s$ are the grasps available at a stable pose $s$.
Given a grasp action $a$ in stable pose $s$, the transition function $T:\mathcal{S} \times \mathcal{A} \times \mathcal{S} \rightarrow [0,1]$ determines the probability distribution over next stable poses. 
The reward function $R:\mathcal{S}\times \mathcal{A} \rightarrow \{0,1\}$ is binary: a grasp is successful and $R(s, a) = 1$ if the grasped object does not fall from the gripper after it is lifted, and $R(s, a) = 0$ otherwise. 
Let $p_{s_a} = \mathbb{E}[R(s, a)]$ be the expected success probability of grasp $a$ on stable pose $s$.
We define a grasping policy as: $\pi: \mathcal{S} \times \mathcal{A} \rightarrow [0, 1]$, where $\pi(a|s)$ denotes the probability of selecting grasp $a$ in pose $s$. We denote the finite horizon of the MDP as $H$. 
The robot initially does not know any of the stable poses or the number of stable poses $N$. If a grasp is successful, the robot 
randomizes the orientation of the object in the gripper, drops the object so that the next stable pose $s'$ is determined by the landing probabilities $\{\lambda_s\}_{s=1}^N$, and records the observed stable pose $s'$. 

We represent the actions, $\mathcal{A}_s$, at each stable pose $s$ as candidate grasps sampled on the object. We use the same method as \citet{Mahlereaau4984} to sample antipodal grasps on each stable pose. We do not make any assumptions on the grasping modality, so in practice these grasps can be sampled from various different grasp planners, including parallel-jaw or suction grasp planners. We denote the number of possible grasps for pose $s$ as $K_s = |\mathcal{A}_s|$ and the total number of grasps over all states as $K = \sum_{s \in \mathcal{S}} K_s$.

An important difference between our problem setting and prior work~\cite{danielczuk2020exploratory} is that we consider settings in which $K$ is large ($>1000$) and thus is of the same order of magnitude as the exploration horizon, $H$. This significantly exacerbates exploration challenges, since there is not enough time to fully explore each grasp, motivating the key innovations in \algabbr.

\textbf{Assumptions:}
In this work, we assume access to the following: (1) a grasp sampler which accepts as input a depth map and outputs a set of candidate grasp configurations on the surface of the depth map with associated robustness values; (2) a robot/gripper that can either execute these grasps or detect that they are in collision; (3) sufficient information in the camera image to detect whether the object stable pose changes; (4) an evaluation function to detect whether a grasp is successful. We note that these assumptions are satisfied by the system we build to instantiate \algabbr in practice. In addition, we make the following assumptions about object's interaction with the environment: (5) if a grasp is unsuccessful, the object either remains in the same stable pose or topples into another stable pose; and (6) there exists a grasp with non-zero success probability on each stable pose. These last two assumptions are consistent with~\cite{danielczuk2020exploratory}. 



\textbf{Metrics:}
We define the \textit{optimality gap}, $\Delta_\pi$ as
\begin{equation}\label{eq:optimality_gap}
    \Delta_\pi = \mathbb{E}_{s\in \mathcal{S}}\left[p^*_s - p_{s_{\pi(s)}}\right] = \sum_{s\in \mathcal{S}} \lambda_s \cdot \left(p^*_s - p_{s_{\pi(s)}}\right),
\end{equation}
where $p_s^* = \max_{a \in \mathcal{A}_s} \mathbb{E}[R(s, a)]$ and $p_{s_{\pi(s)}} = \mathbb{E}[R(s, \pi(s))]$. In simulation, we can evaluate the ground-truth grasp-success probability for a given grasp with robust quasi-static grasp wrench space analysis~\cite{weisz2012pose}. We thus approximate $p_s^*$ by sampling a large number of grasps on each stable pose. Intuitively, the optimality gap $\Delta_\pi$ measures the expected difference, across all stable poses,  between the optimal policy, which selects the best available grasp, and the policy $\pi$. In physical experiments, the optimality gap cannot be computed so we report the grasp-success rate of the learned policy $\pi$.


The objective is to find a policy that minimizes the optimality gap for a given object within $H$ grasp attempts. Denoting a policy learned after $H$ grasp attempts by $\pi_H$, the objective is to identify $\pi_H^*$ such that:
\begin{equation}
    \pi_H^* = \arg\min_{\pi_H} \Delta_{\pi_H}.
\end{equation}

%% file: 4-methods.tex
\section{\algname}
\label{sec:method}
We propose \algname{} (\algabbr{}), a multi-arm bandits algorithm that uses confidence bounds on grasp-success probability to maintain a small active set of candidate grasps. 
\algabbr 
starts with an estimate of the prior success probabilities for all grasps in a large reservoir of possible grasps, and updates their grasp-success probabilities based on online grasp trials using Thompson sampling as in~\citet{danielczuk2020exploratory}. However, unlike BORGES, \algabbr uses the priors and online grasp trials to construct confidence bounds on the grasp-success probabilities for each grasp (Section~\ref{alg:conf_bounds}). 

\algabbr is summarized in Algorithm~\ref{alg:alg}. Once \algabbr visits a stable pose $s$, it checks whether it has visited $s$ (line 4). 
In Sec. \ref{sec:phys-exps}, we describe how to recognize stable poses in the physical setup. If the stable pose $s$ has never been visited (line 5), \algabbr adds the stable pose to the set of visited stable poses $\hat{S}$ (line 6) and initializes an active set of candidate grasps, $\tilde{A}_s$, along with the parameters of a Beta distribution associated with each grasp in the active set (lines 7-8). We rank the grasps in the reservoir by their estimated grasp success probabilities under the Grasp Quality Convolutional Neural Network (GQ-CNN) from Dex-Net 4.0~\cite{Mahlereaau4984} and select the $k=100$ grasps with the highest values. In each iteration, \algabbr executes the grasp with the highest sampled value from the posterior (lines 9-11), observes the outcome (line 12), and updates the posterior distribution~\cite{russo2017tutorial} (lines 13-16). In conjunction, \algabbr also constructs confidence bounds on each of the success probabilities of each grasp (Section~\ref{alg:conf_bounds}). Every $n$ iterations, it uses these confidence bounds to identify and remove the grasps with low robustness (Section~\ref{alg:ci}) (line 18), and replaces them with newly sampled grasps where grasps are ranked by their estimated grasp success probabilities under GQ-CNN (lines 19-20). 

\input{3.75-legs-algorithm} 

\subsection{Constructing Confidence Bounds on Robustness}\label{alg:conf_bounds}
To determine which grasps to remove from the active set, 
\algabbr constructs upper and lower confidence bounds on grasp robustness. We model the success probability of grasp $i$ via $X_i \sim Beta(\alpha_i, \beta_i)$, and empirically select a confidence threshold $\delta$. Then the percent-point function $\texttt{PPF}(X_i, \delta)$, the inverse of the cumulative distribution function $F_{X_i}(x)$, returns the value $x$ such that $F_{X_i}(x) = \delta$. The $(1-\delta)$-lower and -upper confidence bounds for $X_i$ are $X_{i,\ell} = \texttt{PPF}(X_i, \delta)$ and $X_{i,u} = \texttt{PPF}(X_i, 1 - \delta)$, respectively. As a grasp is sampled more often, the interval $[X_{i,\ell}, X_{i,u}]$ tightens to reflect increased certainty in the robustness of the grasp. 

\subsection{Posterior Dependent Grasp Removal} \label{alg:ci}
\algabbr avoids over-exploring less robust grasps by identifying and removing grasps from the active set that are highly likely to be either (1) inferior to another grasp in the active set (\textit{locally suboptimal}) or (2) below a desired global grasp success probability threshold (\textit{globally suboptimal}). Let the highest lower confidence bound across all active grasps be:
$X_\ell^* = \max_{i \in \tilde{A}_s}  X_{i,\ell}$.
We define the set of \textit{locally suboptimal grasps} as the set of grasps for which their $(1-\delta)$-confidence upper bound is worse than the $(1-\delta)$-confidence lower bound for the best grasp in the active set:
\begin{equation}\label{eq:remove_upper_vs_lower}
\mathcal{B}_\ell = \{i:X_{i,u} < X_\ell^*\}.
\end{equation}
Thus, $\mathcal{B}_\ell$ represents the set of grasps that are likely to be inferior to the best known grasp in the active set. However, in the early stages of exploration, we may not yet have sampled a high-performing grasp and $\mathcal{B}_\ell$ may be empty. In these cases, we still desire to remove and resample grasps that, with high-confidence, are clearly low performing. Thus, given a minimum performance threshold $\gamma \in [0, 1]$, we define the set of \textit{globally suboptimal} grasps in the active set (denoted $\mathcal{B}_\gamma$): grasps which have been sampled, but are likely to have success probability less than $\gamma$. We define $\mathcal{B}_\gamma$ as
\begin{equation}\label{eq:remove_gamma}
\mathcal{B}_\gamma = \{i : X_{i,u} < \gamma\}.
\end{equation}
We denote the set of attempted grasps in the active set as $\mathcal{P}$, and let the index of the currently known best grasp be $i*$. The full set of grasps removed by \algabbr is constructed by taking the union of the above sets: $\mathcal{B} = \left(\mathcal{B}_\ell \cup \mathcal{B}_\gamma\right) \cap \mathcal{P} \setminus \{i*\}$. This allows \algabbr to remove grasps which are unlikely to outperform the best known grasp in the current active set.


\subsection{Early Stopping}
\label{alg:early_stopping}
Rather than setting the exploration horizon $H$ to a fixed value, we can set a performance threshold and let \algabbr stop exploring once it has high confidence that it has achieved the desired threshold. This early stopping condition allows \algabbr to efficiently allocate exploration time by only continuing to explore objects that it cannot yet robustly grasp. 

Given a user-specified, minimum performance threshold $\rho_{\min} \in [0,1]$, we want to detect when, with high likelihood, the true performance of \algabbr is above this threshold. More formally, given a confidence parameter $\delta_{\rm stop} \in [0,1]$, we want to calculate a $(1-\delta_{\rm stop})$-confidence lower bound, denoted by $p_\ell$, on the true expected performance of the grasping policy $\pi$,  i.e., we want to find $p_\ell$ such that $Pr \left(p_\ell \leq \mathbb{E}_{s\in S} [p_{s_{\pi(s)}}] \right) \geq 1 - \delta_{\rm stop}$. Then, the robot can stop exploring when $p_\ell \geq \rho_{\rm min}$.

We cannot directly compute $\mathbb{E}_{s\in \mathcal{S}} [p_{s_{\pi(s)}}]$ since we do not know the true stable pose distribution $\mathcal{S}$. Thus, we take a Bayesian approach where we approximate $p_\ell$ by sampling likely values of $\mathbb{E}_{s\in S} [p_{s_{\pi(s)}}]$ given the observed data and then by taking the $\delta_{\rm stop}$-percentile of these samples~\cite{brown2018efficient,brown2018risk}.
First, for each observed stable pose, $s$, we estimate the expected performance of the best grasp as 
    $\hat{p}_s^* = \max_{i \in \mathcal{A}_s} \frac{\alpha_i}{\alpha_i + \beta_i},
    \label{eq:emp_best_arm}$
where $\alpha_i$ and $\beta_i$ are the parameters of the Beta posterior distribution over the success probability of grasp $i$. 
To reason about the performance of \algabbr, we must account for uncertainty over the stable pose distribution, parametrized by the drop probabilities $\lambda_1, \ldots, \lambda_N$. However, $N$ is unknown. Thus, we model our belief over drop probabilities using a Dirichlet posterior distribution over $\hat{N} + 1$ drop probabilities, where $\hat{N}$ is the number of observed stable poses and the +1 allocates probability mass to unobserved stable poses.

Assuming a uniform Dirichlet prior, we take the empirical drop counts $c_1, \ldots, c_{\hat{N}}$ for $\hat{N}$ observed stable poses, and sample from the posterior distribution over stable pose drop probabilities, $Pr( \{ \lambda_s \}_{s=1}^{\hat{N}+1} \mid c_1, \ldots, c_{\hat{N}}, 0)$. 
Due to conjugacy~\cite{diaconis1979conjugate}, the desired posterior distribution is also a Dirichlet distribution with parameters $(\alpha_1 = c_1 + 1, \ldots, \alpha_{\hat{N}} = c_{\hat{N}}+1, \alpha_{\hat{N}+1} = 1)$.
Given a sample, $\{\lambda'_s\}_{s=1}^{\hat{N}+1}$, from the above Dirichlet posterior, we transform it into a sample from the posterior over expected grasp robustness:
    $p'_\pi = \sum_{s=1}^{\hat{N}} \hat{p}_s^* \cdot \lambda'_s$.
where we conservatively assume that the robot will fail to grasp the object in any unseen poses. 
We calculate a $(1-\delta_{\rm stop})$-confidence lower bound on the overall grasp robustness by finding the $\delta_{\rm stop}$ percentile,
    $\hat{p}_\ell = \texttt{PPF}(p'_\pi, \delta_{\rm stop})$,
using $M$ samples of $p'_\pi$. 

%% file: 3.75-legs-algorithm.tex
\begin{algorithm}[t]
\vspace{2pt}
\SetAlgoLined
    \textbf{Input:} object $o$, grasp sampler $f_\theta$, resample interval $n$, number of active grasps $k$ \\
    Initialize the set of visited stable poses $\hat{S} = \varnothing$ \\
    
    \For{$t=1, 2, \ldots$}{
        Recognize the current stable pose $s$ \\
        \uIf{$s \notin \hat{S}$}{
            $\hat{S} \leftarrow \hat{S} \cup \{s\} $\\
            Use $f_\theta$ to sample $k$ grasps as the active set $\tilde{A}_s$\\
            Set $\alpha_{i}$ and $\beta_i$ based on prior for all $i \in \tilde{A}_s$\\
        }
        \ForEach{grasp $i \in \tilde A_s$}{
            sample $\phi_i \sim Beta(\alpha_i+1, \beta_i+1)$
        }
        Execute grasp $i = \text{argmax}_{j \in \tilde A_s} \phi_j$ \\
        Observe reward $r = R(s,i)$ \\
        \uIf{$r=1$}{
            $\alpha_{i} \leftarrow \alpha_{i} + 1$
        }
        \Else{
            $\beta_{i} \leftarrow \beta_{i} + 1$
        }
        \uIf{$t \equiv 0 \pmod n$} {
            Remove the grasps in $\mathcal{B} = \mathcal{B}_\ell \cup \mathcal{B}_\gamma$ from $\tilde{A}_s$ (see equations \eqref{eq:remove_upper_vs_lower} and \eqref{eq:remove_gamma})\\
            Sample $|\mathcal{B}|$ new grasps using $f_{\theta}$ \\
            For each new grasp $j=1,\ldots|\mathcal{B}|$, set $\alpha_{j},\beta_{j}$ using prior from $f_\theta$ and add new grasp to $\tilde{A}_s$ \\
        }
    }
    \caption{\algname (\algabbr)}
     \label{alg:alg}
\end{algorithm}

%% file: 5-experiment-and-result.tex
\section{Simulation Experiments} \label{sec:experiments_and_results} 
\subsection{Experimental Setup}
\label{subsec:sim-setup}
We first evaluate \algabbr in Exploratory Grasping with a variety of adversarial objects in simulation. Same as in Danielczuk et al.~\cite{danielczuk2020exploratory}, we consider 14 Dex-Net 2.0 Adversarial objects~\cite{mahler2017dex} and all 39 EGAD!\ Adversarial evaluation objects~\cite{morrison2020egad}. We use Dex-Net 4.0~\cite{Mahlereaau4984} to sample a large reservoir of $K=2000$ grasps for each stable pose. We also use GQ-CNN to set the Beta prior for \algabbr following the method from~\cite{li2020,danielczuk2020exploratory}.
Using the method outlined in Section~\ref{sec:method}, we update the active grasp set after every $n=100$ timesteps and use $\delta=0.05$ for constructing grasp confidence intervals with upper confidence threshold $\gamma = 0.2$. All experiments use a time horizon of $H=3000$. We run 10 trials of each algorithm with 10 rollouts per trial, where each trial involves sampling a different reservoir of grasps, and each rollout for a trial involves running a grasp exploration algorithm.
\subsection{Baselines}
\label{subsec:baselines}
We compare \algabbr against five baseline algorithms: Dex-Net, Tabular Q-Learning (TQL), BORGES ($K_s=100$), BORGES ($K_s=2000$), and \algabbr(-AS). Dex-Net greedily chooses the best grasp evaluated by Dex-Net 4.0~\cite{Mahlereaau4984} for each stable pose and does not do any online exploration. BORGES ($K_s=100$) leverages a prior calculated by GQ-CNN to seed grasp success probability estimates, and then performs Thompson Sampling for each encountered stable pose to explore an initial active set of $100$ grasps sampled on each of the poses. While BORGES ($K_s=100$) is provided with the same initial active set as \algabbr, unlike \algabbr, BORGES ($K_s=100$) does not update its set over time. However, different from~\cite{danielczuk2020exploratory}, it is not guaranteed that there will exist successful grasps on all stable poses when $K_s=100$. This implies that BORGES ($K_s=100$) may not be able to transit between stable poses. The $K_s=100$ Upper Bound refers to the optimality gap if on each stable pose, the best grasp in the active set is selected. BORGES ($K_s=2000$) is identical to BORGES ($K_s=100$), but instead directly explores the full reservoir of $K_s=2000$ sampled grasps. TQL implements tabular Q-learning  on the full reservoir of $K_s = 2000$ sampled grasps where each pose is a separate state $s$ and each action $a$ is a grasp on that pose and a Q-table $Q[s, a]$ is constructed to keep track of the corresponding 1-step Q-values. 
The values in the $Q$-table are initialized using the GQ-CNN prior and actions are chosen based on an $\epsilon$-greedy policy~\cite{sutton2018reinforcement} with $\epsilon=0.1$.  Finally, \algabbr(-AS) is not provided with an initial active set, but instead operates on the full reservoir of $K_s=2000$ grasps and uses the posterior dependent removal procedure in Section~\ref{alg:ci} to remove grasps from the reservoir. 

\begin{table*}
    \vspace{4pt}
    \centering
    \begin{tabular}{lp{1.8cm}p{1.8cm}p{1.8cm}p{1.8cm}p{1.8cm}p{1.8cm}p{1.8cm}}
        \toprule
         Dataset & Dex-Net & TQL & \makecell{BORGES\\($K_s=100$)} & \makecell{$K_s=100$\\Upper Bound} & \makecell{BORGES\\($K_s=2000$)} & \algabbr(-AS) & \algabbr \\
         \midrule
            Dex-Net & $0.56 \pm 0.07$ & $0.23 \pm 0.08$ & $0.13 \pm 0.07$ & $0.08 \pm 0.04$ & $\mathbf{0.04 \pm 0.03}$ & $0.22 \pm 0.06$ & $\mathbf{0.04 \pm 0.03}$\\
            EGAD! & $0.59 \pm 0.03$ & $0.32 \pm 0.04$ & $0.25 \pm 0.04$ & $0.13 \pm 0.03$ & $0.17 \pm 0.03$  & $0.28 \pm 0.04$ & $\mathbf{0.14 \pm 0.03}$\\
        \bottomrule
    \end{tabular}
    \caption{\textbf{Grasping in Simulation Aggregated Results: }We show the optimality gap (mean $\pm$ standard error) achieved by \algabbr and baselines after $H=3000$ steps of exploration averaged over the objects in the Dex-Net Adversarial and EGAD! evaluation datasets. \algabbr achieves a lower optimality gap than all baselines, indicating that \algabbr is able to discover new high-performing grasps.}
    \label{tab:sim-grasping-aggregate}
\end{table*}


\begin{figure*}
    \centering
    \includegraphics[width=0.93\linewidth]{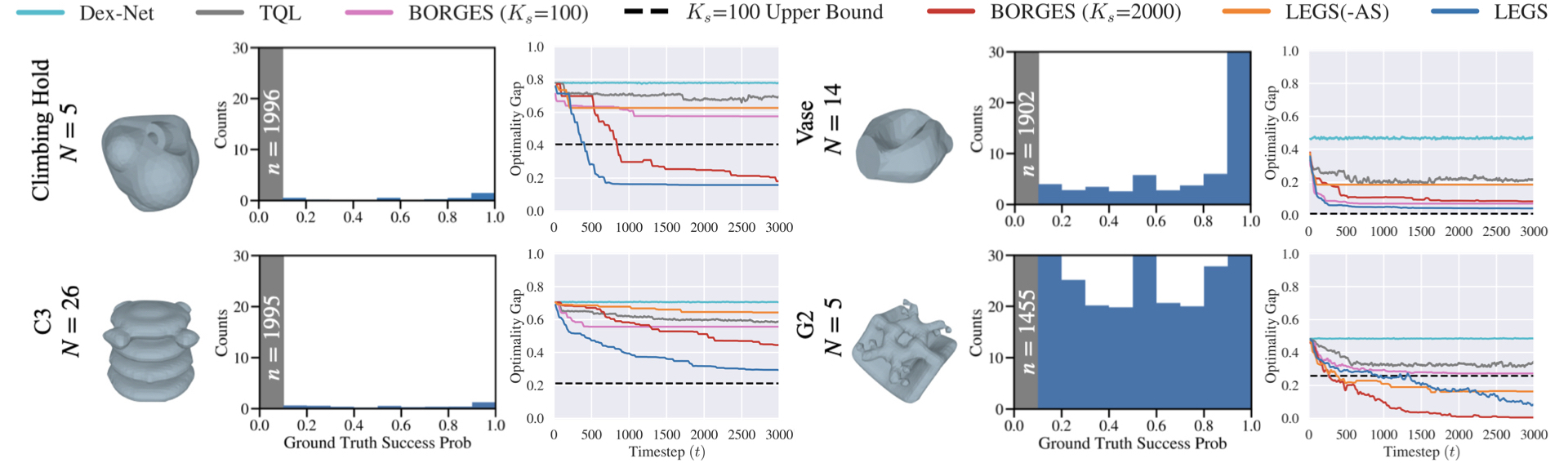}
    \caption{\textbf{Simulated Grasping Experiments Case Study: } We report the performance of \algabbr and baselines on four specific objects to investigate how object properties affect performance. For each object, we include a 3D rendering of the object and the number $N$ of stable poses (left), a histogram of the ground truth grasp success probabilities over 2000 sampled grasps (middle), and learning curves (right). 
    }
    \label{fig:sim_analysis}
\end{figure*}

\subsection{Experimental Results}
\label{subsec:sim-results}
We first study aggregated results of \algabbr and baselines over objects in the Dex-Net Adversarial and EGAD! evaluation datasets in Table~\ref{tab:sim-grasping-aggregate}. 
We find that \algabbr performs better than or equal to the baseline algorithms on 10 out of 14 objects in the Dex-Net Adversarial dataset, and on 25 out of 39 objects in the EGAD! evaluation dataset. In comparison, the best performing baseline algorithm, BORGES ($K_s=2000$), only performs at least as well as rest of the algorithms on 5 out of 14 Dex-Net Adversarial objects and 14 out of 39 EGAD! evaluation dataset. 
On all of these objects we find that Dex-Net, which is not updated online, has high optimality gap, motivating online grasp exploration. 
The improvement for \algabbr over \algabbr(-AS) and BORGES ($K_s = 2000$) indicates the increased efficiency of restricting exploration to a small active set, while the gap between \algabbr and BORGES ($K_s = 100$) indicates the importance of updating this active set over time to prune poor performing grasps while discovering new, high-quality grasps outside of the initial active set. BORGES ($K_s = 100$) cannot outperform the success rate of the best grasp in its initial set ($K_s=100$ upper bound). By contrast, \algabbr, retains the efficiency of only exploring a small set of grasps while also being able to adapt this set over time to obtain successful grasps on difficult-to-grasp stable poses and reach a lower optimality gap. TQL learns much more slowly than BORGES because it fails to leverage the structure in the grasp exploration problem and does not learn separate policies for each stable pose.

In Figure~\ref{fig:sim_analysis}, we study \algabbr and baselines on specific objects. We show two objects (Climbing Hold and C3) where \algabbr converges faster to high performing grasps than prior algorithms and two objects (F6 and Turbine Housing) where \algabbr does not outperform all baselines. We find that when high performing grasps are abundant, \algabbr may converge to suboptimal grasps. However, when there are only few successful grasps, \algabbr can converge to good grasps much faster than baselines. If high quality grasps are already in the active set, \algabbr can rapidly distinguish them from other grasps. If the active set does not contain successful grasps, \algabbr can quickly replace bad grasps in the active set.


\subsection{Early Stopping Results}
\label{subsec:early-stopping}
Next, we study the accuracy and effectiveness of the early stopping criterion (Section~\ref{alg:early_stopping}). We test the proposed high-confidence performance bound across all objects in the Dex-Net Adversarial object set (individual results per object are reported in the supplement). We check whether \algabbr has reached the stopping condition every 100 grasps for a horizon of $H=3000$ total grasp attempts and use $\delta_{\rm stop} = 0.05$, resulting in a 95\%-confidence lower bound $\hat{p}_\ell$. We sample $M=3000$ samples to estimate $\hat{p}_\ell$.  

We first test how often the predicted bound is a true lower bound on performance. We find that, on average, across all Dex-Net Adversarial objects, our empirical lower bound is a 95.8\%-accurate lower bound on the true performance over the true stable pose distribution. Thus, $\hat{p}_\ell$ forms an empirically valid $(1-\delta_{\rm stop})$-confidence lower bound. We next test the tightness of our lower bound. On average, the difference between the true performance of \algabbr and our empirical lower bound is only 2.97\%. These results suggest that our lower bound is highly accurate and tight enough to provide a practical signal for when the robot can safely stop exploring. 

\begin{figure}
    \vspace{2pt}
    \centering
    \includegraphics[width=\linewidth]{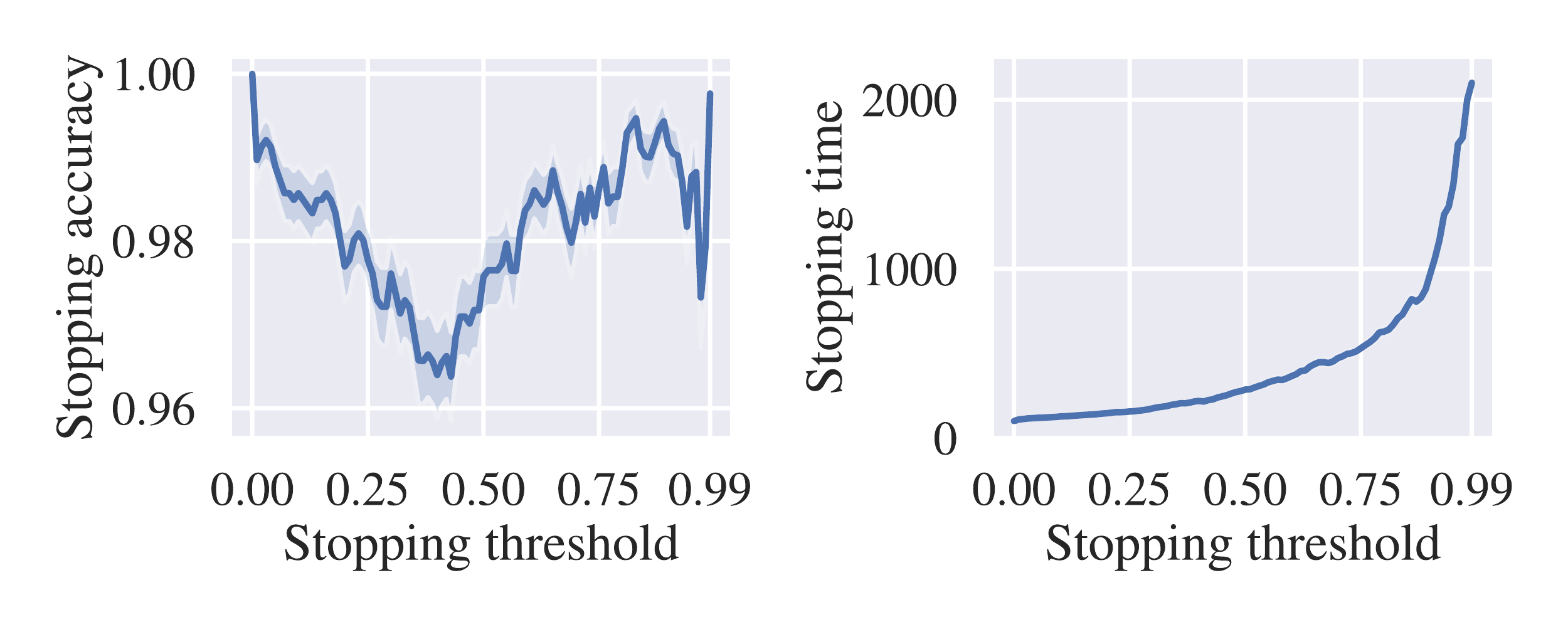}
    \caption{\textbf{Early Stopping Threshold Sensitivity: } We evaluate early stopping over the Dex-Net Adversarial object set in simulation with a range of stopping thresholds, $\rho_{\rm min}$. We use a 95\%-confidence lower bound on expected grasp robustness. \textbf{Left:} We plot the accuracy averaged over all objects and find that our empirical lower bound (Section~\ref{subsec:early-stopping}) is highly accurate across all stopping thresholds, $\rho_{\rm min}$. \textbf{Right:} We plot the number of steps before stopping, averaged across all objects. Intuitively, the required exploration time increases with higher performance thresholds. Importantly, the average number of steps before stopping is much lower than the 3000-step horizon.}
    \label{fig:early_stopping_sensitivity}
\end{figure}

We next study, in simulation, the use our high-confidence bounds on performance for early stopping. As described in Section~\ref{alg:early_stopping}, given a user-specified, minimum performance threshold, $\rho_{\min}$, the robot stops exploring when the lower confidence bound $\hat{p}_\ell$ is greater than $\rho_{\min}$. When the robot chooses to stop exploring the object, we evaluate the ground truth performance of the learned policy and evaluate whether the true performance is also above the threshold $\rho_{\min}$. We evaluate a wide range of thresholds and plot the results in Figure~\ref{fig:early_stopping_sensitivity}. Results suggest that we can achieve highly accurate early stopping, allowing the robot to accurately terminate exploration well before the full horizon of $3000$ steps.


%% file: 6-physical-experiments.tex
\section{Physical Experiments}
\label{sec:phys-exps}
In this section, we discuss our experimental setup for physical experiments, the methods we used to enable intervention-free grasp exploration on a physical robot and results evaluating the performance of \algabbr and BORGES ($K_s=2000$) across 3 physical objects.

\subsection{Experimental Setup} \label{subsec:phys-setup}
To deploy exploratory grasping algorithms on a physical robot, we modify the perception system introduced in~\citet{danielczuk2020exploratory} to sample grasps and identify changes in the object stable pose. We capture a depth image of the object from an overhead camera, deproject it into a point cloud using the known camera intrinsics, demean the point cloud, and apply 3600 evenly spaced rotations to the point cloud around the camera's optical axis. We measure the chamfer distance between the rotated point clouds with previously cached point clouds and find the pair of point clouds that serves as the closest match. As in~\citet{danielczuk2020exploratory}, if at least 80\,\% of the points are less than 0.02\,mm away from the closest points in the cached point cloud, we classify the two point clouds as belonging to the same stable pose. If none of the cached point clouds satisfies this condition, the point cloud is cached and treated as a new stable pose. If there exists a matching point cloud, we further align the translation and rotation of the point cloud via iterative closest point~\cite{chetverikov2002trimmed}. 

Upon discovery of a new stable pose, we use Dex-Net 4.0~\cite{Mahlereaau4984} to sample, evaluate, and cache grasps in the grasp reservoir. Thus, \algabbr can explore grasps on objects with unknown geometries and unknown numbers of stable poses. 

\subsection{Self-Supervised Exploratory Grasping with \algabbr}
\label{subsec:self-sup}
\citet{danielczuk2020exploratory} find that re-dropping the object during experiments often cause it to fall out of the workspace, requiring extensive human effort to reset the object. To enable the robot to collect grasp data without human intervention, we introduce strategies to prevent the object from toppling out of the workspace while maintaining access to a wide variety of grasps. We drop the object within a bowl (Fig.\ref{fig:splash}), where the object's rebound height is lower than the rim of the bowl. The bowl allows the object to stay in the visible range of the overhead camera. However, the bowl's rim can be an obstacle to grasps. We introduce two autonomous \emph{reset} behaviors to address this: (1) we center the object above the bowl before dropping the object, ad (2) when the object topples near the boundary, the robot pushes the object towards the center of the bowl to improve grasp access~\cite{danielczuk2018linear}. 


\subsection{Experimental Results}
\label{subsec:results}

\begin{figure}[t] 
    \vspace{2pt}
    \centering
    \includegraphics[width=\linewidth]{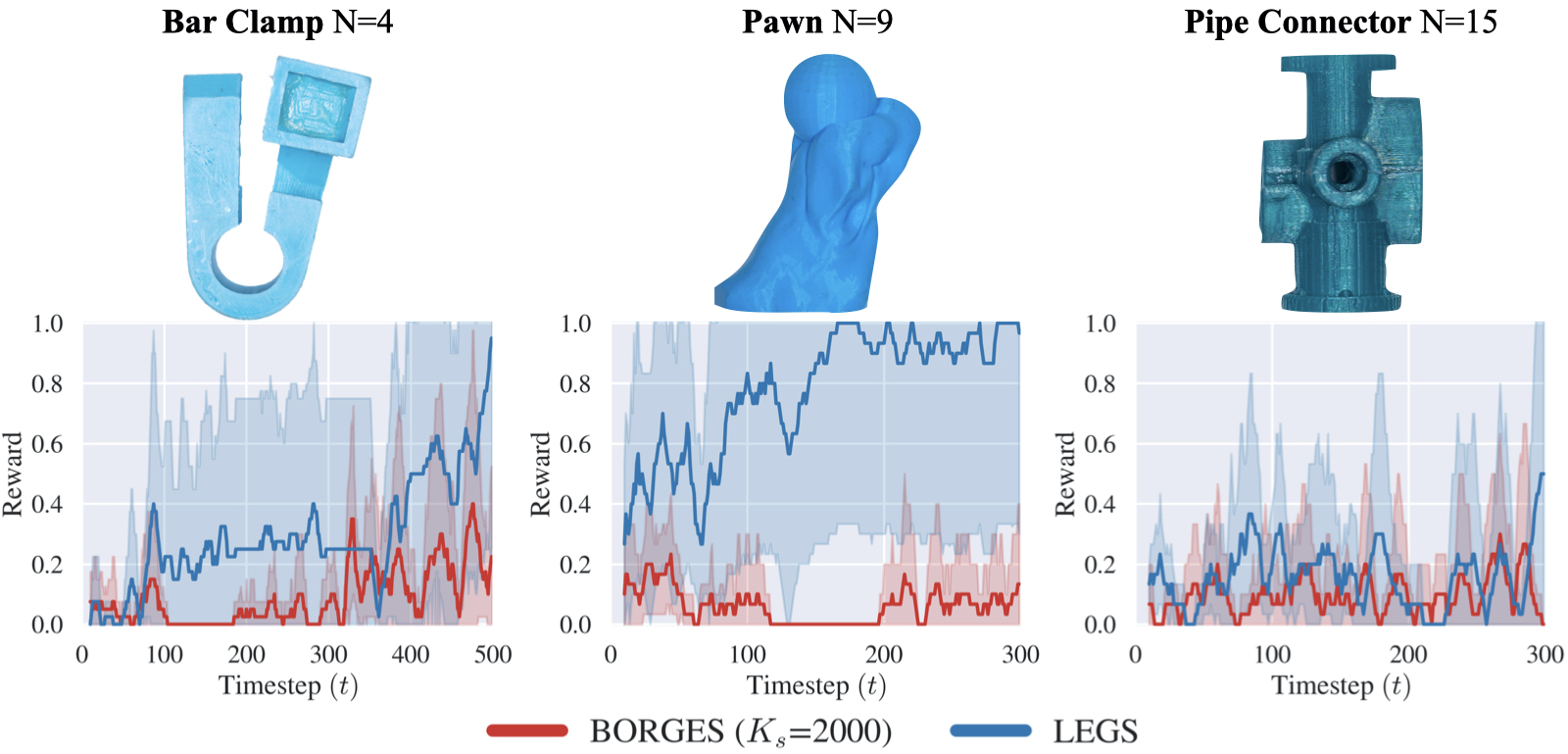}
    \caption{\textbf{Physical Experiments Results:} We compare \algabbr with BORGES ($K_s=2000$) on three objects (Bar Clamp, Pawn, and Pipe Connector) from the Dex-Net Adversarial Dataset \cite{mahler2017dex} in physical experiments. All physical experiments are completed within 3 hours. \algabbr significantly outperforms BORGES ($K_s=2000$) on Bar Clamp and Pawn, with minor improvement on Pipe Connector.}
    \label{fig:physical_results}
\end{figure}

Figure \ref{fig:physical_results} shows learning curves from physical experiments comparing \algabbr with BORGES ($K_s=2000$) on three challenging objects from the Dex-Net Adversarial Dataset \cite{mahler2017dex}. We run 3 trials with 1 rollout per trial for each object. We find that on 2 out of the 3 objects, \algabbr is able to outperform BORGES ($K_s=2000$) and identify high-performing grasps within a few hundred timesteps of online exploration.

%% file: 7-summary.tex
\section{Discussion}
We present \algname, an algorithm which efficiently explores large sets of grasps by adaptively constructing a small active set of promising grasps. Experiments suggest that \algabbr identifies high-performing grasps more efficiently than baseline algorithms across 53 objects in simulation experiments and on three challenging objects in physical trials. We also propose a novel early stopping condition by computing a high-confidence lower bound on the expected grasp performance. Simulation results suggest that this high-confidence lower bound is highly accurate and tight. In future work, we will analyze \algabbr to determine how the quality of the Dex-Net prior and the distribution over grasp success probabilities affect its convergence rate. Moreover, we will search for possible ways for \algabbr to generalize across different stable poses and objects.

%% file: 8-acknowledgement.tex
\small{\section{Acknowledgments}
This research was partially supported by Google, Siemens, Toyota Research Institute,
the Berkeley AI Research (BAIR) Lab and by equipment grants from PhotoNeo and NVIDIA. 
Any opinions, findings, and conclusions or recommendations expressed are those of the author(s) and do not necessarily reflect the views of the sponsors. Michael Danielczuk is supported by the National Science Foundation Graduate Research Fellowship Program under Grant No. 1752814. We thank Brijen Thananjeyan and Ellen Novoseller, who provided helpful feedback, and Adam Lau for his masterful photos.}

%% file: 9-appendix.tex
\onecolumn
\section*{Appendix} \label{sec:appendix}
We select the hyperparameters based on the ablation studies done on the \emph{Bar Clamp} object in the Dex-Net 2.0 adversarial objects~\cite{mahler2017dex}. Specifically, we perform ablation studies on two hyperparameters: $s$ the strength of GQCNN prior and $\delta$ for constructing grasp confidence intervals. For each ablation experiments, we run 10 trials of each algorithm with 10 rollouts per trial, where each trial involves sampling a different reservoir of grasps, and each rollout for a trial involves running a grasp exploration algorithm. All experiments are run over a time horizon of $H=3000$. Our experiments (Table \ref{tab:sim-grasping-aggregate}) show that this set of hyperparameters tuned on a single object can be applied across many objects.

\subsection{Sensitivity to Prior Strength}
The effect of the strength of GQCNN prior is first studied in~\citet{li2020}. In~\citet{danielczuk2020exploratory}, $s$ is set to 5. Our sensitivity experiments show that when more grasps are sampled on each stable pose, $s=1$ shows the best result. In this set of experiments, $\delta = 0.07$.

\begin{table}[h]
    \vspace{4pt}
    \centering
    \begin{tabular}{cc}
        \toprule
         Strength of Prior & Bar Clamp \\
         \midrule
            $s$ = 0 & $0.285 \pm 0.055$ \\
            $s = \mathbf{1}$ & $\mathbf{0.125 \pm 0.036}$ \\
            $s$ = 2 & $0.128 \pm 0.038$ \\
            $s$ = 3 & $0.171 \pm 0.041$ \\
            $s$ = 4 & $0.148 \pm 0.036$ \\
            $s$ = 5 & $0.187 \pm 0.042$ \\
        \bottomrule
    \end{tabular}
    \caption{\textbf{Sensitivity to GQCNN prior strength $s$:} We show the optimality gap (mean $\pm$ standard error). }
    \label{tab:ablation-s}
\end{table}

\subsection{Sensitivity to Confidence Interval Parameter}
We also performed sensitivity experiments on $\delta$ for constructing confidence intervals for all grasps. Intuitively, using a small $\delta$ will lead to a larger confidence interval, which may slow down the updates to the active set. Using a large $\delta$ will lead to a smaller confidence interval, which may lead to false positives when identifying grasps with low success rate. In our ablation experiments, we find that $\delta = 0.05$ gives the best performance on the bar clamp object. In this set of experiments, $s = 1$.

\begin{table}[h]
    \vspace{4pt}
    \centering
    \begin{tabular}{cc}
        \toprule
         Confidence Parameter & Bar Clamp \\
         \midrule
            $\delta = 0.01$ & $0.129 \pm 0.040$ \\
            $\delta = \mathbf{0.05}$ & $\mathbf{0.103 \pm 0.031}$ \\
            $\delta = 0.10$ & $0.156 \pm 0.043$ \\
            $\delta = 0.15$ & $0.148 \pm 0.041$ \\
            $\delta = 0.20$ & $0.214 \pm 0.051$ \\
            $\delta = 0.25$ & $0.174 \pm 0.044$ \\
        \bottomrule
    \end{tabular}
    \caption{\textbf{Sensitivity to confidence interval parameter $\delta$:} We show the optimality gap (mean $\pm$ standard error).}
    \label{tab:ablation-delta}
\end{table}